\documentclass{article}

\usepackage[main,preprint]{neurips_2026}
\makeatletter
\renewcommand{\@notice}{}
\makeatother

\usepackage[utf8]{inputenc}
\usepackage[T1]{fontenc}
\usepackage{hyperref}
\usepackage{url}
\usepackage{amsfonts}
\usepackage{nicefrac}
\usepackage{microtype}
\usepackage{graphicx}
\usepackage{booktabs}
\usepackage{multirow}
\usepackage{subcaption}
\usepackage{amsmath}
\usepackage{amssymb}
\usepackage{float}
\usepackage{xcolor}

\definecolor{compositionalorange}{HTML}{B24C00}

\title{Looped Language Models \\ Improve Compositional Tool Calling}

\author{%
  \begin{tabular}{ccc}
    Andrei Cristian Popescu &
    Haitz Sáez de Ocáriz Borde &
    Pietro Liò \\
    \texttt{acp96@cam.ac.uk} &
    \texttt{hs788@cam.ac.uk} &
    \texttt{pl219@cam.ac.uk}
  \end{tabular}
  \\[0.6em]
  Department of Computer Science and Technology \\
  University of Cambridge, United Kingdom
}

\begin{document}

\maketitle

\begin{abstract}
    Looped language models have shown promising results on reasoning benchmarks, yet their potential for agentic tool use remains largely unexplored. We study this question in compositional tool-calling settings, where models must coordinate multiple API calls, maintain intermediate state, and preserve dependencies across tool interactions. We evaluate native and retrofitted looped language models on API-Bank, BFCL, and NESTful, comparing looped and non-looped models trained under matched supervised fine-tuning recipes and varying recurrent depth at inference time. In controlled experiments, recurrent computation generally benefits compositional and dependency-aware tool use, while providing smaller and more model-dependent gains on isolated API invocation. Accuracy on multi-step tool use generally increases with recurrent depth; adaptive inference, however, achieves a more favorable compute-performance trade-off by allocating additional computation only when needed. Our results suggest that looped language models are a promising architecture for agentic systems that require reliable planning, coordination, and execution of compositional tool use workflows.
\end{abstract}

\begin{center}
    \vspace{-0.3em}

    \begin{minipage}[t]{0.46\textwidth}
        \centering
        \includegraphics[
            width=0.9\linewidth
        ]{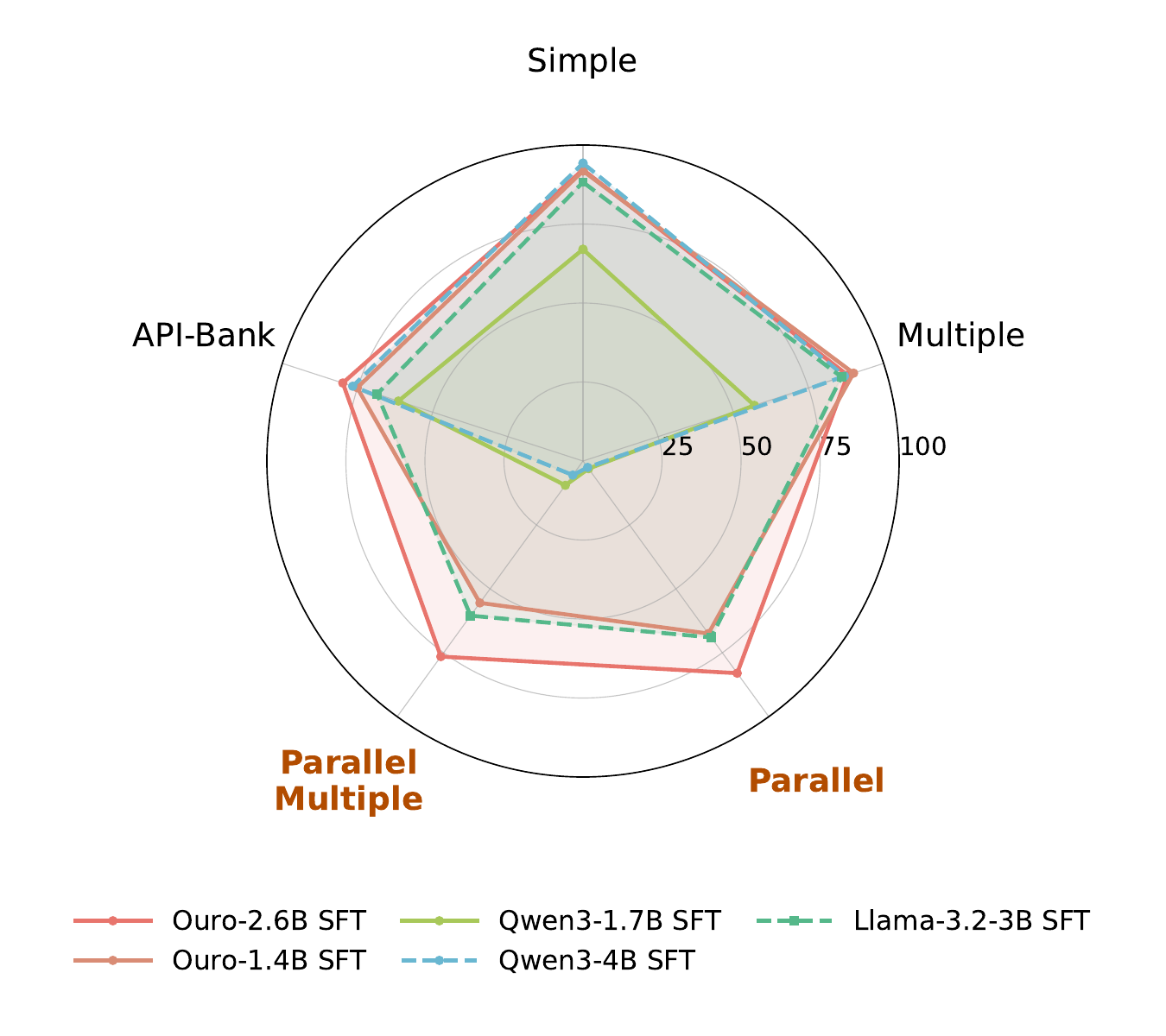}

        \vspace{-0.2em}
        {\small (a) Native looped and baseline models.}
    \end{minipage}
    \hfill
    \begin{minipage}[t]{0.46\textwidth}
        \centering
        \includegraphics[
            width=0.9\linewidth
        ]{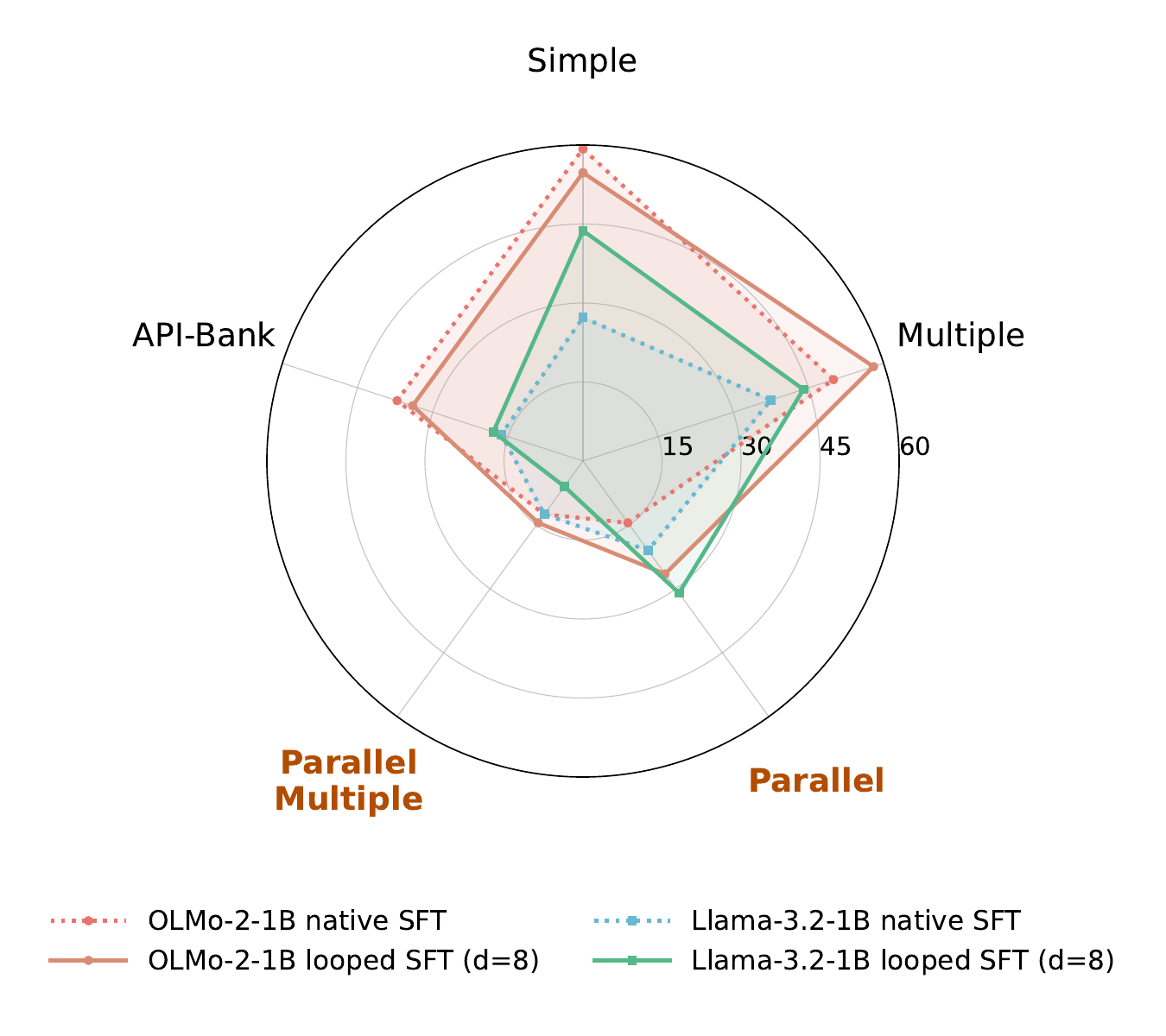}

        \vspace{-0.2em}
        {\small (b) Retrofitted and native backbones.}
    \end{minipage}

    \vspace{0.2em}

    \captionsetup{hypcap=false}
    
    \captionof{figure}{
        \textbf{Looped computation primarily improves compositional tool use.}
        \textbf{(a)} Under matched supervised fine-tuning conditions, Ouro models show
        their largest relative advantages on multi-call BFCL categories, while
        differences on the predominantly single-call API-Bank benchmark are less
        consistent.
        \textbf{(b)} Retrofitting recurrence into Llama-3.2-1B and OLMo-2-1B likewise
        improves several compositional categories relative to their non-recurrent
        counterparts.
        Category labels are color-coded: compositional tasks are shown in \textcolor{compositionalorange}{orange},
        while non-compositional tasks are shown in black.
    }
    \label{fig:headline}
\end{center}

\section{Introduction}

Pretrained language models are increasingly used as decision-making components in agentic systems, selecting and invoking external tools to complete user-directed tasks \citep{mrkl,react,toolformer,hugginggpt}. Tool use provides a controlled setting for studying how pretrained representations are translated into actions: while simple requests may require only a single function call, more complex tasks require models to compose multiple calls, maintain intermediate state, and coordinate sequential or parallel dependencies. Reliable tool use therefore requires more than producing an individually plausible invocation; models must construct and preserve structured action sequences across multiple decisions.

Looped language models provide a natural mechanism for strengthening this form of structured decision making. Instead of relying on a single forward pass, they repeatedly refine latent representations before generating each token, increasing test-time computation without increasing parameter count. Although this iterative computation has shown promising results on reasoning benchmarks, its benefits for agentic behavior remain underexplored.

\noindent\textbf{Our main contributions are:}
\begin{enumerate}
\item We assess whether iterative latent computation improves compositional tool use in a controlled setting requiring structured, multi-step decision making.
\item Our evaluation covers native and retrofitted looped models on API-Bank, BFCL, and NESTful, with matched fine-tuning comparisons and shared-backbone Llama and OLMo retrofits.
\item The largest benefits appear on multi-call and dependency-aware tasks, suggesting that recurrent computation is particularly useful when action selection must preserve structure across multiple decisions.
\item Increasing recurrent depth generally improves compositional tool use, while adaptive inference achieves a better compute-performance trade-off by allocating fewer iterations when additional refinement is unnecessary.
\end{enumerate}

\section{Related Work}

We review two lines of work most closely related to our study: recurrent-depth architectures for latent computation, and methods for planning and composition in tool-using language models.

\paragraph{Recurrent Depth and Latent Computation} Looped Transformers increase effective model depth by repeatedly applying a shared Transformer block, decoupling test-time computation from parameter count. Early recurrent and parameter-sharing architectures include the Universal Transformer \citep{UT}, Deep Equilibrium Models \citep{bai2019deepequilibriummodels}, and ALBERT \citep{albert}. More recent work introduced looped Transformers as a method for latent reasoning, where recurrent iterations refine hidden representations rather than extending the context with explicit reasoning tokens \citep{reasoninglatentthoughtspower,Huginn,Ouro,loopformer,coconut,zelikman2024quietstarlanguagemodelsteach,huang2025fastquietstarthinkingthought}. Looped models have demonstrated strong performance on algorithmic reasoning, length generalization, adaptive computation, and latent test-time scaling \citep{loopedtransformersprogrammablecomputers,anil2022exploringlengthgeneralizationlarge,yang2024loopedtransformersbetterlearning,kohli2026loopthinkgeneralize,popescu2026adaptivedepth}. We test whether these capabilities transfer to tool-use settings.

\paragraph{Planning and Composition in Tool-Using Models}
Tool-augmented language models extend autoregressive generation by invoking external functions for retrieval, computation, and interaction with software environments. Early work demonstrated that language models can interleave reasoning with tool execution \citep{react}, learn API usage from self-supervised annotations \citep{toolformer}, or route requests to specialized neural and symbolic modules \citep{mrkl}. More recent approaches improve compositional tool use through explicit planning, including decomposing tasks into executable workflows \citep{hugginggpt,chameleon}, retrieving relevant APIs from large tool collections \citep{toollm,gorilla}, and constructing dependency-aware execution graphs that schedule sequential and parallel function calls \citep{llmcompiler}. Unlike these approaches, which primarily enhance tool use through external planning, retrieval, or execution strategies, our work investigates whether additional latent iterative computation in looped Transformers benefits compositional tool calling. Benchmarks such as API-Bank \citep{apibank}, BFCL \citep{bfcl}, and NESTful \citep{nestful} are used to evaluate increasingly complex tool-use capabilities, ranging from single-function invocation to parallel, sequential, and nested API compositions.

\section{Background}
\label{sec:background_tool_calling}

We first introduce the recurrent computation used by looped Transformers, and then formalize compositional tool calling as structured prediction over sets of tool invocations and their dependencies.

\paragraph{Looped Transformers} In contrast to classical Transformers that stack distinct Transformer layers, looped Transformers repeatedly apply a shared block of Transformer layers over multiple recurrent iterations. Each iteration refines the latent representation, increasing inference-time computation while keeping the parameter count fixed. Let $h^{(0)}$ denote the initial hidden representation of an input sequence. At recurrent iteration $t$, the shared Transformer block updates the latent state according to \( h^{(t)} = F_{\theta}\!\left(h^{(t-1)}\right), \) where $F_{\theta}$ denotes the shared recurrent block. A shared output head then produces a prediction after every recurrent iteration, \( \pi_\theta^{(t)}(y \mid x)=g_{\theta}\!\left(h^{(t)}\right), \) producing a sequence of refined predictions $\{\pi_\theta^{(1)},\ldots,\pi_\theta^{(T)}\}$, where $T$ is the maximum number of recurrent iterations. Looped Transformers naturally enable adaptive computation. Rather than executing a fixed number of recurrent iterations for every token, looped Transformers can dynamically allocate computation by selecting an exit iteration using either a learned halting policy or post-hoc criteria \citep{pondernet,Ouro,popescu2026adaptivedepth}. During inference, easier predictions can terminate after fewer recurrent iterations, improving the cost–accuracy trade-off.

\paragraph{Compositional Tool Use Formalism} Let $\mathcal{T}=\{\tau_1,\ldots,\tau_N\}$ denote the tools available to a language model, where each tool $\tau_i$ is specified by a function name, a natural-language description, and an argument schema. Given a user request $x$, the language model predicts a distribution over structured tool calls \( p_\theta(c_k \mid x), \) where \( c_k=(f_k,a_k), \) $f_k\in\mathcal{T}$ is the selected function and $a_k$ denotes its instantiated arguments. Executing $c_k$ returns an observation $o_k$, which may be incorporated into subsequent model predictions. A tool use solution can be represented as a directed acyclic graph \( G_x=(C_x,E_x), \) where $C_x=\{c_1,\ldots,c_K\}$ is the set of tool calls and $(c_i,c_j)\in E_x$ indicates that call $c_j$ depends on the result of call $c_i$. A single-call task has $|C_x|=1$. Independent calls have no dependencies and may be executed in parallel, whereas sequential calls are constrained by an ordering. In a dependent or nested sequence, the observation returned by an earlier call is used to instantiate an argument of a later call, such that \( a_j = \phi_j(x,o_{i_1},\ldots,o_{i_m}), \) for predecessor calls $\{c_{i_1},\ldots,c_{i_m}\}$. More complex workflows may combine parallel branches with sequential dependencies. We refer to the construction of multi-call solutions as compositional tool calling. Under this representation, independent calls have \(E_x=\varnothing\) and dependent workflows have \(E_x\neq\varnothing\). Thus, \(C_x\) specifies the required tool calls, while \(E_x\) specifies the output-to-input dependencies between them. Using this formulation, existing benchmarks emphasize different components of the tool-calling process. API-Bank primarily evaluates individual tool calls $c_k=(f_k,a_k)$, focusing on correct tool selection, argument grounding, and API invocation. BFCL evaluates both single-call function selection and independent multi-call generation. Its Simple and Multiple categories require \(|C_x|=1\), whereas its Parallel and Parallel-Multiple categories require \(|C_x|>1\), with \(E_x=\varnothing\) throughout. Finally, NESTful emphasizes the dependency structure $E_x$, where outputs returned by earlier calls are consumed as arguments of later calls, forming hierarchical execution chains. These benchmarks evaluate node-level call prediction, structured multi-call composition, and dependency-aware execution.

\section{Experimental Setup}
\label{sec:experimental_setup}

This section describes the models, training procedure, evaluation benchmarks, and inference protocols used in our experiments. We compare looped and non-looped models under matched supervised fine-tuning recipes, and evaluate both fixed-depth and adaptive looped inference to study the role of further computation during tool use.

\subsection{Model Families and Variants}
We evaluate looped language models in two complementary settings. First, we compare the looped Ouro-1.4B and Ouro-2.6B models against standard Transformer baselines from the Qwen3 and Llama families at similar parameter scales. We fine-tune the corresponding base checkpoints using the same dataset and optimization settings, and separately evaluate released instruction-tuned Qwen and Llama checkpoints up to 8B parameters as stronger public reference systems. The larger instruction-tuned checkpoints also provide reference points whose single-pass inference cost is closer to that of multiple loop iterations. Second, we evaluate retrofitted looped models based on OLMo-2-1B and Llama-3.2-1B. For each model, we compare the looped checkpoint against its corresponding non-looped parent after supervised fine-tuning with the same data and optimization recipe. These comparisons more directly isolate the effect of introducing looped computation while preserving the underlying pretraining family. Architectural details for Ouro and the retrofitted models are provided in Appendix~\ref{app:models}.

\paragraph{Supervised Fine-Tuning} All controlled models are supervised fine-tuned on the Hermes function calling dataset. Training examples are formatted using ChatML with the Hermes JSON tool-calling protocol. More details about the dataset can be found in Appendix~\ref{app:hermes_dataset}. All experiments use a maximum sequence length of 4,096 tokens, a 90/10 train-validation split, two training epochs, AdamW with learning rate $2\times10^{-5}$, cosine learning-rate decay with 3\% warmup, gradient accumulation of 2, per-device batch size of 2, LoRA rank 32, and bf16 precision. The experiments were run on an NVIDIA A100 80GB GPU. The Ouro models are fine-tuned using the original training objective,
\[ \mathcal{L}(\theta,\phi) = \sum_{t=1}^{T} p_{\phi}(t \mid x)\, \mathcal{L}_{\theta}^{(t)}(x) + \beta\, \mathrm{KL} \!\left( p_{\phi}(\cdot \mid x) \,\|\,\pi(\cdot) \right), \]
where $\beta=0.1$, $\mathcal{L}_{\theta}^{(t)}$ denotes the next-token cross-entropy at loop iteration $t$, $p_{\phi}(t\mid x)$ is the learned distribution over loop iterations, and $\pi$ is the uniform prior over loop depths. The retrofitted looped models do not learn an exit distribution. Instead, the number of loops $r$ is sampled independently for each training batch using the Poisson-lognormal depth distribution proposed in the original paper \cite{retrofittedrecurrence}. The model is supervised at the final readout after the sampled number of iterations:
\[
\mathcal{L}(\theta)=\mathbb{E}_{x}\mathbb{E}_{r\sim\mathcal{D}_{\mathrm{PLN}}}\Bigg[\sum_{\ell=1}^{M-1}\Bigg(-\log p_{\theta}^{(r)}\!\left(x_{\ell+1}\mid x_{1:\ell}\right)+\beta\,\mathrm{KL}\!\left(p_{\theta_0}(\cdot\mid x_{1:\ell})\,\middle\|\,p_{\theta}^{(r)}(\cdot\mid x_{1:\ell})\right)\Bigg)\Bigg].
\]
Here, $\beta=0.1$, $p_{\theta}^{(r)}$ is the next-token distribution of the adapted model after $r$ loop iterations and $p_{\theta_0}$ is the corresponding distribution of the frozen pre-adaptation model. Unlike Ouro, which learns a per-token distribution over loop iterations, the retrofitted model is trained using randomly sampled recurrence depths and always computes the supervised loss from the final recurrent readout.

\subsection{Evaluation Benchmarks}

We evaluate tool use performance using three benchmarks covering different aspects of function calling: BFCL v3, NESTful, and API-Bank. These benchmarks aim to assess isolated API invocation, compositional multi-tool reasoning, and hierarchical tool execution. Additional details and representative examples for all three benchmarks are provided in Appendix~\ref{app:benchmark_examples}.

\paragraph{BFCL v3.} BFCL v3 \citep{bfcl} is our primary evaluation benchmark. We evaluate using the official BFCL v3 single-turn protocol and report AST accuracy on the non-live benchmark split. Following the benchmark protocol, we report results separately for the Simple, Multiple, Parallel, and Parallel-Multiple categories. Simple tasks require one tool call. Multiple tasks also require a single call, but the model must select the correct function from several candidate tool definitions. Parallel tasks require multiple independent invocations of a single function. Parallel-Multiple tasks combine function selection with multiple independent invocations across several candidate functions. Thus, in BFCL, Multiple refers to the number of candidate tools rather than the number of generated calls. Only the Parallel and Parallel-Multiple categories require multiple calls. The evaluator standardizes tool calls before comparison, making the evaluation robust to formatting differences while requiring semantic equivalence of the predicted calls. We report the official AST accuracy, which scores a prediction as correct if its standardized abstract syntax tree matches the reference tool call. Representative examples for each category are provided in Appendix~\ref{app:bfcl_examples}.

\paragraph{NESTful.} NESTful \citep{nestful} evaluates hierarchical tool use through nested API workflows, where later calls depend on outputs produced by earlier ones. Unlike BFCL, the benchmark emphasizes long-range dependencies and multi-step execution rather than isolated function prediction. Following the benchmark protocol, we report Win Rate as the primary execution-based metric, together with function, parameter, partial-sequence, and full-sequence scores. A representative nested execution example is provided in Appendix~\ref{app:nestful_examples}.

\paragraph{API-Bank.} API-Bank \citep{apibank} evaluates tool use across a diverse collection of real-world APIs spanning multiple domains. In contrast to BFCL and NESTful, most tasks involve a single API invocation, putting more emphasis on selecting the appropriate API and generating correct arguments than on coordinating multiple interacting tool calls. Hence, it complements the compositional reasoning benchmarks by measuring API grounding and invocation accuracy. We report three complementary metrics. Call Correctness applies the released API-specific functional checks, Exact requires an exact match of the API name and complete argument dictionary after parsing and conservative normalization, and Parse is the fraction of outputs that can be parsed as valid API calls. A representative API-Bank instance is provided in Appendix~\ref{app:apibank_examples}.

\subsection{Analyzing the Effect of Iterative Computation} 
\label{res:depth}
We evaluate the effect of successive recurrent iterations using both fixed-depth and adaptive inference protocols. These experiments are performed on the BFCL v3 evaluation split described above and a 500-example subset of NESTful. The NESTful subset is selected by taking examples evenly across the benchmark's official ordering before any experiments are run, and is held constant across all models and inference settings.

\paragraph{Fixed-Depth Inference.} To evaluate the effect of additional looped computation, we perform fixed-depth inference by executing a predetermined number of loop iterations for every generated token. For the Ouro models, we evaluate depths 1 to 4, corresponding to the model's maximum recurrent depth. For the retrofitted recurrent Llama-3.2-1B and OLMo-2-1B models, we evaluate 1, 2, 4, and 8 loop iterations. Modifying only the inference depth while keeping model parameters fixed isolates the contribution of recurrence independently of training.

\paragraph{Adaptive Inference.} In addition to fixed-depth inference, we evaluate adaptive looped computation using Ouro's pretrained exit gate. During generation, each token exits the inference loop once the cumulative exit probability exceeds a confidence threshold $q$. We evaluate thresholds $q\in\{0.1,0.3,0.5,0.7,0.8\}$ and report both task performance and the average selected loop depth, computed as the mean number of loop iterations used across all generated tokens.

\section{Results}

In this section, we start by evaluating whether looped models improve tool use over non-looped baselines. We then investigate whether these improvements arise from iterative computation itself through controlled inference-time depth ablations and whether recurrent computation can be allocated adaptively to improve the compute–performance trade-off.

\begin{table}[t]
\centering
\caption{
BFCL semantic AST correctness (\%; higher is better).
Overall is the aggregate score across all task categories.
The lower block compares looped and non-looped variants of each backbone,
fine-tuned using identical data and optimization settings;
looped variants use a fixed inference depth of 8 loop iterations.
Base checkpoints omitted from the lower block score near zero throughout.
}
\label{tab:bfcl}
\footnotesize
\setlength{\tabcolsep}{4.5pt}
\renewcommand{\arraystretch}{0.98}
\begin{tabular}{llccccc}
\toprule
Model & Training
& Simple
& Multiple
& Parallel
& Par.-Mult.
& Overall \\
\midrule
Ouro-1.4B
& Base
& 65.3 & 67.0 & 35.0 & 44.5 & 55.4 \\
& SFT
& 91.8 & 90.0 & 67.5 & 55.5 & 79.3 \\
\cmidrule(lr){1-7}
Ouro-2.6B
& Base
& 80.5 & 75.5 & 2.0 & 1.0 & 47.9 \\
& SFT
& 92.3 & 88.0 & 83.0 & 76.5 & 86.4 \\
\specialrule{\heavyrulewidth}{\aboverulesep}{\belowrulesep}
Qwen3-1.7B
& Base
& 1.5 & 0.0 & 0.0 & 0.0 & 0.6 \\
& SFT
& 67.0 & 57.0 & 3.0 & 9.5 & 40.7 \\
& Instruct
& 91.8 & 91.5 & 83.5 & 81.0 & 87.9 \\
\cmidrule(lr){1-7}
Qwen3-4B
& Base
& 72.0 & 57.5 & 1.0 & 0.5 & 40.6 \\
& SFT
& 94.3 & 87.0 & 2.5 & 5.5 & 56.7 \\
& Instruct
& 93.0 & 92.5 & 87.5 & 88.5 & 90.9 \\
\cmidrule(lr){1-7}
Qwen3-8B
& Instruct
& 95.5 & 96.0 & 91.5 & 89.5 & 93.6 \\
\specialrule{\heavyrulewidth}{\aboverulesep}{\belowrulesep}
Llama-3.2-1B
& Instruct
& 18.2 & 2.5 & 4.0 & 4.5 & 9.5 \\
\cmidrule(lr){1-7}
Llama-3.2-3B
& SFT
& 88.2 & 86.0 & 69.0 & 60.5 & 78.4 \\
& Instruct
& 33.8 & 37.0 & 0.0 & 0.0 & 20.9 \\
\cmidrule(lr){1-7}
Llama-3.1-8B
& Instruct
& 46.5 & 41.0 & 0.0 & 0.0 & 26.8 \\
\specialrule{\heavyrulewidth}{\aboverulesep}{\belowrulesep}
OLMo-2-1B
& SFT
& 59.2 & 50.0 & 14.5 & 12.5 & 39.1 \\
& Looped SFT
& 55.0 & 58.0 & 26.5 & 14.5 & 41.8 \\
\cmidrule(lr){1-7}
Llama-3.2-1B
& SFT
& 29.8 & 28.5 & 14.0 & 5.0 & 21.4 \\
& Looped SFT
& 43.5 & 40.5 & 31.0 & 6.0 & 32.9 \\
\bottomrule
\end{tabular}
\end{table}

\paragraph{Looped language models improve tool use.} Tables~\ref{tab:bfcl}, \ref{tab:nestful}, and \ref{tab:apibank} report results on BFCL, NESTful, and API-Bank. The controlled comparisons use identical datasets, optimization schedules, and LoRA configurations for looped and non-looped models. We evaluate both native Ouro models against similarly sized Transformer baselines and retrofitted recurrent variants of Llama-3.2-1B and OLMo-2-1B against their non-recurrent parents. Publicly released instruction-tuned Qwen and Llama checkpoints are included as reference systems.

\begin{table}[t]
\centering
\caption{
NESTful official evaluation (higher is better).
Win Rate is the primary metric.
}
\label{tab:nestful}
\footnotesize
\setlength{\tabcolsep}{6pt}
\renewcommand{\arraystretch}{1.08}
\begin{tabular}{llccccc}
\toprule
Model & Training
& Function F1
& Parameter F1
& Partial
& Full
& Win Rate \\
\midrule

Ouro-1.4B
& Base
& 0.905 & 0.539 & 0.149 & 0.091 & 0.110 \\
& SFT
& 0.899 & 0.566 & 0.219 & 0.131 & 0.191 \\
\cmidrule(lr){1-7}

Ouro-2.6B
& Base
& 0.920 & 0.595 & 0.207 & 0.128 & 0.190 \\
& SFT
& 0.922 & 0.680 & 0.295 & 0.204 & 0.371 \\
\midrule

Qwen3-1.7B
& Base
& 0.000 & 0.000 & 0.000 & 0.000 & 0.000 \\
& SFT
& 0.000 & 0.000 & 0.000 & 0.000 & 0.000 \\
& Instruct
& 0.924 & 0.555 & 0.202 & 0.109 & 0.134 \\
\cmidrule(lr){1-7}

Qwen3-4B
& Base
& 0.000 & 0.000 & 0.000 & 0.000 & 0.000 \\
& SFT
& 0.458 & 0.309 & 0.156 & 0.001 & 0.063 \\
& Instruct
& 0.971 & 0.703 & 0.285 & 0.196 & 0.292 \\
\cmidrule(lr){1-7}

Qwen3-8B
& Instruct
& 0.979 & 0.774 & 0.329 & 0.246 & 0.345 \\
\midrule

Llama-3.2-3B
& Base
& 0.911 & 0.561 & 0.208 & 0.155 & 0.146 \\
& SFT
& 0.911 & 0.495 & 0.175 & 0.093 & 0.095 \\
& Instruct
& 0.929 & 0.419 & 0.160 & 0.033 & 0.060 \\
\cmidrule(lr){1-7}

Llama-3.1-8B
& Instruct
& 0.657 & 0.342 & 0.137 & 0.030 & 0.073 \\

\bottomrule
\end{tabular}
\end{table}

\begin{table}[t]
\centering
\caption{
\textbf{API-Bank evaluation} (\%; higher is better).
\textbf{(a)} Controlled comparison between looped and non-looped models trained using identical supervised fine-tuning recipes. The lower block pairs each retrofitted backbone with its non-looped counterpart; looped variants are evaluated at a fixed inference depth of 8 loop iterations.
\textbf{(b)} Comparison against publicly released instruction-tuned checkpoints. Call Correctness measures functional API-call correctness, Exact requires an exact API-name-and-argument match, and Parse reports valid call generation.
}
\label{tab:apibank}
\footnotesize
\renewcommand{\arraystretch}{1.08}
\begin{tabular}[t]{@{}c@{}}
\textbf{(a) Controlled models} \\[0.35em]
\setlength{\tabcolsep}{4.5pt}
\begin{tabular}{lcccccc}
\toprule
& \multicolumn{3}{c}{Base}
& \multicolumn{3}{c}{SFT} \\
\cmidrule(lr){2-4}
\cmidrule(lr){5-7}
Model
& Call & Exact & Parse
& Call & Exact & Parse \\
\midrule
Ouro-1.4B
&73.0&67.6&95.1
&75.1&70.2&97.2\\
Ouro-2.6B
&79.2&76.9&99.2
&79.9&77.1&99.7\\
\midrule
Qwen3-1.7B
&5.1&5.1&13.9
&61.4&57.8&93.6\\
Qwen3-4B
&71.2&71.2&99.2
&76.6&73.8&99.7\\
\midrule
Llama-3.2-1B
&1.7&1.3&10.8
&16.3&14.1&55.3\\
Llama-3.2-3B
&0.3&0.3&0.3
&68.5&64.5&99.7\\
\specialrule{\heavyrulewidth}{\aboverulesep}{\belowrulesep}
OLMo-2-1B
&1.9&0.5&35.7
&37.1&33.2&99.5\\
OLMo-2-1B (Loop)
&0.1&0.0&1.5
&34.0&30.3&90.7\\
\cmidrule(lr){1-7}
Llama-3.2-1B
&1.7&1.3&10.8
&16.3&14.1&55.3\\
Llama-3.2-1B (Loop)
&0.1&0.0&1.5
&17.9&16.2&43.4\\
\bottomrule
\end{tabular}
\end{tabular}
\hfill
\begin{tabular}[t]{@{}c@{}}
\textbf{(b) Released instruct checkpoints} \\[0.35em]
\setlength{\tabcolsep}{5pt}
\begin{tabular}{lccc}
\toprule
Model
& Call
& Exact
& Parse \\
\midrule
Ouro-1.4B (SFT)
&75.1&70.2&97.2\\
Ouro-2.6B (SFT)
&79.9&77.1&99.7\\
\midrule
Qwen3-1.7B
&79.4&74.0&100.0\\
Qwen3-4B
&78.7&75.3&100.0\\
Qwen3-8B
&79.9&76.6&100.0\\
\midrule
Llama-3.2-1B
&5.1&5.1&13.9\\
Llama-3.2-3B
&74.6&70.7&98.5\\
Llama-3.1-8B
&78.8&74.8&100.0\\
\bottomrule
\end{tabular}
\end{tabular}
\end{table}

In both experimental settings, looped models perform better on compositional tool-calling tasks, whereas gains on isolated API invocation are smaller and model-dependent. On BFCL, the smallest differences occur on Simple tasks, where both looped and non-looped SFT models already perform strongly. Larger gains appear on the compositional categories, particularly Parallel and Parallel-Multiple. NESTful shows a similar pattern, where the Ouro models improve on the full evaluation, while the fixed-depth subset indicates that additional recurrence can also benefit the retrofitted models on hierarchical workflows. API-Bank shows much smaller differences, consistent with its greater emphasis on individual API selection and argument generation. Despite their relatively small parameter counts, the Ouro models remain competitive with several released Qwen and Llama instruction checkpoints.

\begin{figure}[t]
    \centering
    \includegraphics[width=0.9\textwidth]{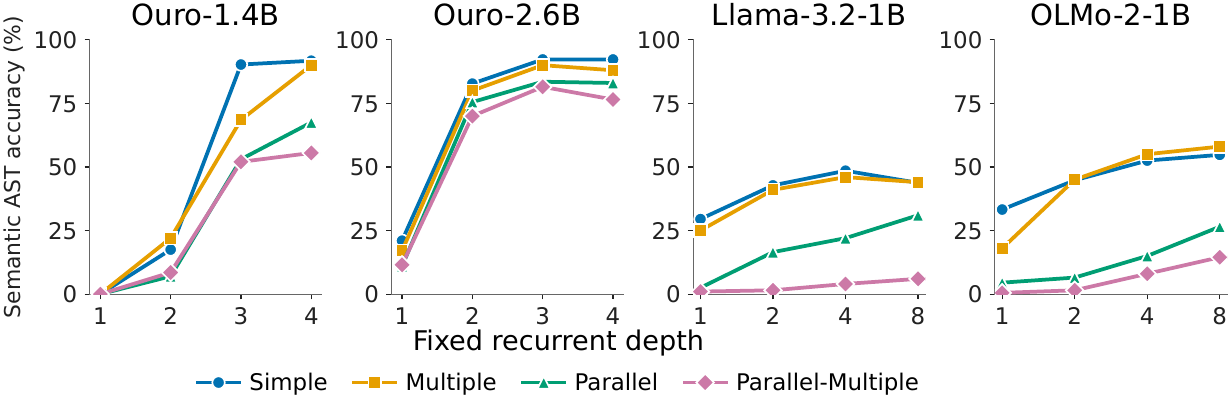}
    \caption{
    BFCL semantic AST accuracy by task category as fixed recurrent depth increases.
    Each curve reports accuracy on Simple, Multiple, Parallel, and Parallel-Multiple tasks.
    }
    \label{fig:bfcl_fixed_depth_subtasks}
\end{figure}

\paragraph{Tool-use performance increases with recurrent depth.} The results in the previous section demonstrate that looped language models outperform their non-looped counterparts, but they do not distinguish improvements arising from the loops themselves from those due to training. To isolate the contribution of looped computation, we vary the number of inference loop iterations while keeping the model fixed. Figure~\ref{fig:bfcl_fixed_depth_subtasks} shows that BFCL accuracy generally rises with recurrent depth, with the largest gains on compositional categories. For Ouro-1.4B and the retrofitted Llama model, Simple tasks saturate earlier, whereas the tool-selection-heavy Multiple category and the multi-call Parallel and Parallel-Multiple categories continue to benefit from further computation before plateauing. OLMo-2-1B shows a similar pattern on multi-call tasks. Ouro-2.6B instead reaches near-saturation after three iterations in all categories, suggesting that the larger model requires fewer refinement steps.

\begin{figure}[!htpb]
    \centering

    \begin{subfigure}[t]{0.45\textwidth}
        \centering
        \includegraphics[width=0.82\linewidth]{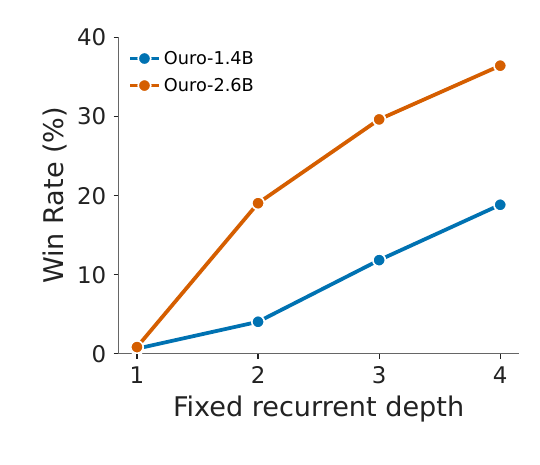}
        \caption{Native recurrent Ouro models.}
    \end{subfigure}
    \hspace{0.02\textwidth}
    \begin{subfigure}[t]{0.45\textwidth}
        \centering
        \includegraphics[width=0.82\linewidth]{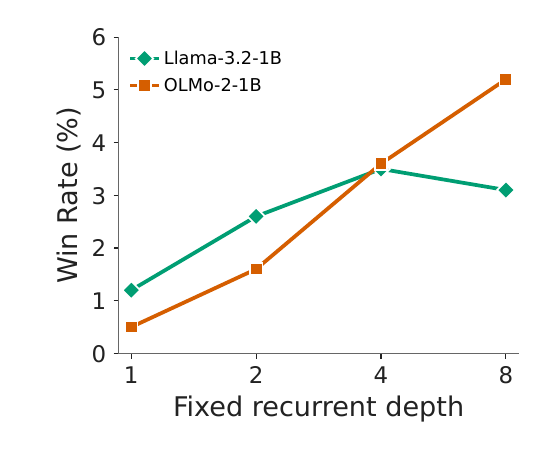}
        \caption{Retrofitted recurrent Llama model.}
    \end{subfigure}

    \caption{
    NESTful Win Rate as fixed recurrent depth increases. Native recurrent Ouro
    models benefit from additional recurrent computation, whereas the retrofitted
    recurrent Llama baseline remains substantially weaker.
    }
    \label{fig:nestful_fixed_depth}
\end{figure}

NESTful exhibits the same depth effect: Win Rate increases as the number of recurrent iterations grows (Figure~\ref{fig:nestful_fixed_depth}). Since later function calls explicitly depend on outputs from previous calls, the improvement with depth indicates that recurrence benefits dependent tool interactions. Across both Ouro models and the retrofitted Llama model, the effect persists when only inference depth is varied, separating it from differences introduced during training.

\begin{figure}[!htpb]
    \centering

    \begin{subfigure}[t]{0.49\textwidth}
        \centering
        \includegraphics[width=1.0\linewidth]{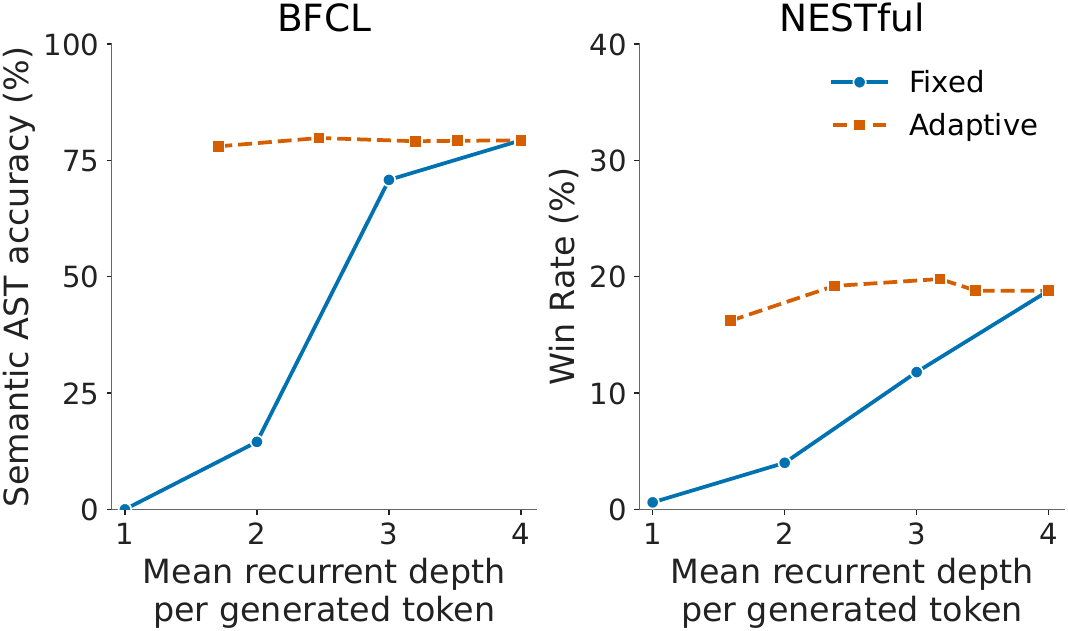}
        \caption{Ouro-1.4B.}
    \end{subfigure}
    \hfill
    \begin{subfigure}[t]{0.49\textwidth}
        \centering
        \includegraphics[width=1.0\linewidth]{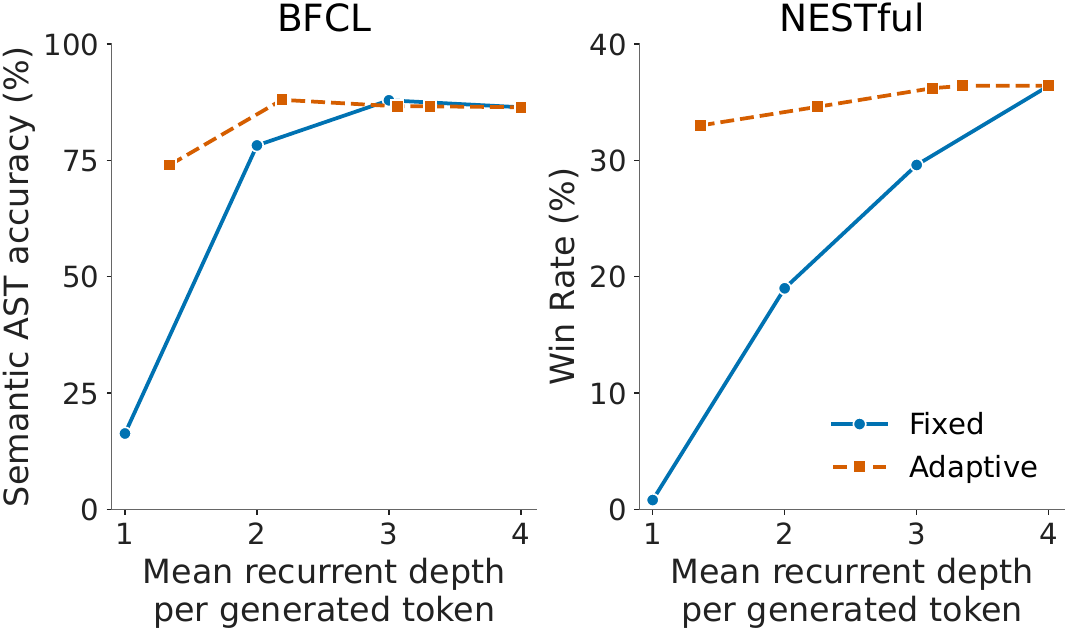}
        \caption{Ouro-2.6B.}
    \end{subfigure}

    \caption{
    Adaptive recurrent computation on BFCL and NESTful.
    The x-axis reports the mean recurrent depth per generated token. Adaptive stopping improves the compute-performance frontier by allocating additional recurrent iterations only when beneficial. On BFCL, adaptive stopping matches or slightly exceeds the best fixed-depth operating point while executing fewer recurrent iterations on average. On NESTful, Ouro-2.6B reaches the same Win Rate as fixed depth 4 while using fewer recurrent iterations per generated token.
    }
    \label{fig:adaptive_compute}
\end{figure}

\paragraph{Adaptive looped computation increases tool-calling efficiency.} The previous subsection shows that additional inference loop iterations improve tool use performance but eventually saturate, with the saturation point varying across models and tasks. This suggests that a fixed inference depth is suboptimal. A key advantage of looped language models is that they naturally support adaptive computation, allowing the model to dynamically determine how much iterative refinement each prediction requires. Next, we evaluate adaptive looped computation, where the model exits the inference loop on a per-token basis using the Ouro adaptive exit gate.

Figure~\ref{fig:adaptive_compute} compares adaptive and fixed-depth inference on BFCL and NESTful. Adaptive inference yields a better performance–compute trade-off for both Ouro models. It recovers most of the gains from deeper recurrence while using fewer loop iterations per generated token, and in several settings matches or exceeds the best fixed-depth configuration at lower average cost. Allocating recurrent computation by token difficulty thus yields a better compute–performance trade-off than applying a uniform depth.

\begin{figure}[!htpb]
    \centering
    \includegraphics[width=0.9\textwidth]{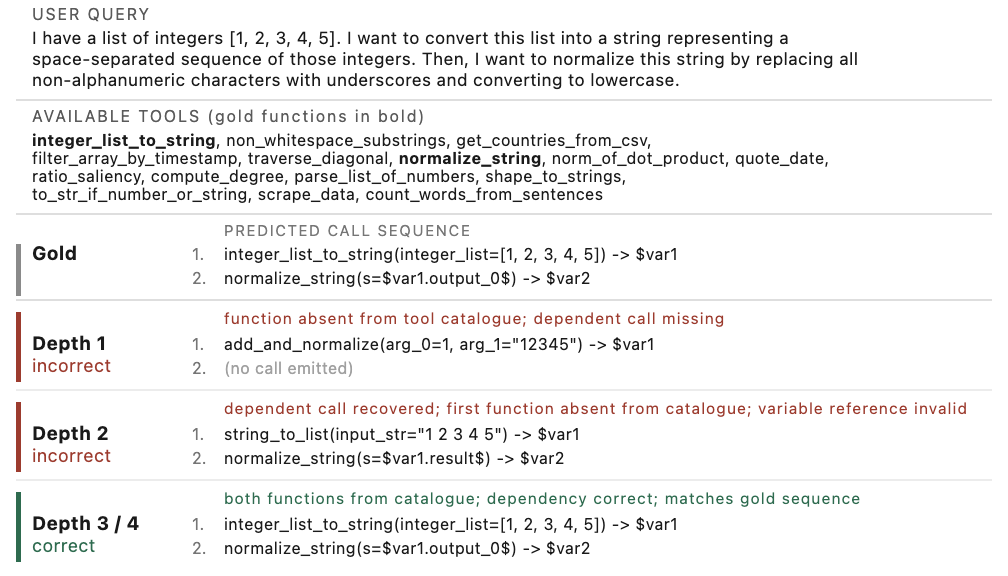}
    \caption{\textbf{Iterative refinement across recurrent depths.} Ouro-1.4B on a
    two-step NESTful composition task. Calls are shown as
    \texttt{function(argument=value) -> answer}. At depth 1 the model emits a single call to a
    function absent from the tool catalogue and omits the dependent call; at depth 2
    it recovers the two-call structure but the first function is again absent from
    the catalogue and the variable reference is invalid. Depths 3 and 4 match the
    gold sequence, including the output-to-input reference
    \texttt{\$var1.output\_0\$}.}
    \label{fig:qualitative_refinement}
\end{figure}

\paragraph{Iterative computation refines tool invocations.} Figure~\ref{fig:qualitative_refinement} shows a representative NESTful example. At shallow recurrent depths, the model omits the dependent call or produces invalid function and variable references. By depth 3, it recovers the complete tool sequence and the correct output-to-input dependency, after which the prediction remains unchanged. This example suggests that recurrence can correct semantic errors in call structure and dependency binding, rather than merely output formatting. Additional examples appear in Appendix~\ref{app:qualitative-examples}.

\section{Conclusion}

Our empirical evaluation shows that looped computation is particularly beneficial when tool calling requires composition, dependency tracking, or coordination across multiple calls. On BFCL and NESTful, native Ouro models and retrofitted Llama and OLMo models generally outperform comparable non-looped baselines on structured tool use tasks. Notably, although post-hoc recurrence improves several compositional tasks, retrofitted models remain substantially weaker than natively recurrent models on deeply nested workflows, suggesting that the effectiveness of recurrent refinement may depend on how representations are shaped during pretraining. By contrast, gains are smaller and more model-dependent on API-Bank, where most examples involve isolated API invocation. Fixed-depth experiments further show that multi-step tool use performance generally increases with recurrent depth, and qualitative analysis indicates that successive iterations can incrementally refine function selection, call structure, and intermediate dependencies. Adaptive stopping recovers most of the gains from deeper recurrent inference using fewer loop iterations on average, yielding a better compute-performance trade-off. The results presented in this paper suggest that looped language models are a promising foundation for agentic systems that must dynamically allocate computation while planning, coordinating, and executing compositional tool use workflows.

\section{Limitations}

Our conclusions are drawn from a deliberately controlled setting. For the Ouro models in particular, no non-looped counterpart trained under identical pretraining conditions is publicly available, so we approximate a matched comparison by evaluating against both the Qwen3 and Llama families under the same fine-tuning recipe. The retrofit experiments, which share a backbone with their non-recurrent parents, isolate the architectural change more directly. Because additional recurrent iterations increase per-token compute, we also include larger instruction-tuned checkpoints as reference points, with inference costs that approximately upper-bound those of the looped models. Finally, all three benchmarks are static, single-turn evaluations. Extending this analysis to live, multi-turn settings such as the BFCL live and multi-turn categories, where the model must recover from failed executions across an episode, is left to future work.

\bibliographystyle{plainnat}
\bibliography{main}

\clearpage
\appendix

\section{Looped Models Architectural Details}
\label{app:models}

This section summarizes the architectures and training procedures of the looped language models evaluated throughout this work. We first describe the native recurrent Ouro models and then the retrofitted recurrent models, focusing on the architectural and training differences relevant to our evaluation.

\subsection{Ouro}
\label{app:ouro}

Ouro~\citep{Ouro} is a family of pretrained looped language models that performs iterative latent computation by repeatedly applying a shared stack of $L$ decoder-only Transformer layers. The recurrent block consists of standard Transformer layers with multi-head self-attention, Rotary Position Embeddings (RoPE), SwiGLU feed-forward networks, and sandwich RMSNorm. Rather than stacking independently parameterized layers, the same recurrent block is applied for up to $T_{\max}=4$ recurrent iterations, allowing additional test-time computation without increasing the number of model parameters. The released checkpoints are Ouro-1.4B, containing a 24-layer recurrent stack, and Ouro-2.6B, obtained by doubling the recurrent stack to 48 layers and continuing pretraining. Both models use a hidden size of 2048 and are pretrained on approximately 7.7T tokens.

At each recurrent iteration $t$, Ouro predicts both the next-token distribution and a conditional halting probability \( \lambda^{(t)}(x)=\sigma\!\left(\mathrm{Linear}_{\phi}(h^{(t)})\right), \) which induces an exit distribution over recurrent depths
\( p_{\phi}(t\mid x) = \lambda^{(t)}(x) \prod_{j<t} \left(1-\lambda^{(j)}(x)\right), \) with the remaining probability mass assigned to the final recurrent iteration. This exit distribution is used both during training to weight the prediction losses across recurrent depths and during inference to determine the adaptive computation depth.

Training proceeds in two stages. Stage I jointly optimizes the language model and halting gate using
\[ \mathcal{L} = \sum_{t=1}^{T_{\max}} p_{\phi}(t\mid x)\, \mathcal{L}^{(t)} - \beta H\!\left(p_{\phi}(\cdot\mid x)\right), \]
where $\mathcal{L}^{(t)}$ is the language-model loss after recurrent iteration $t$. The entropy regularization term is equivalent to a KL penalty towards a uniform prior over exit depths, encouraging the model to utilize multiple recurrent iterations before specializing the halting policy.

In Stage II, the language-model backbone is frozen and only the halting gate is optimized. Rather than supervising language-model predictions, the gate is trained from the marginal utility of executing one additional recurrent iteration. Given the detached improvement in prediction loss \( I_i^{(t)} = \max\!\left( 0, \mathcal{L}^{(t-1)}_{i,\mathrm{stop}} - \mathcal{L}^{(t)}_{i,\mathrm{stop}} \right), \) the improvement score is converted into a soft exit target $y_i^{(t)}$, and the gate is optimized using the binary cross-entropy objective
\[ \mathcal{L}_{\mathrm{gate}} = -\sum_{t=1}^{T_{\max}} \Big[ y^{(t)}\log \lambda^{(t)} + (1-y^{(t)}) \log\!\left(1-\lambda^{(t)}\right) \Big]. \]
This second stage improves the calibration of the adaptive exit policy while leaving the pretrained recurrent representations unchanged.

Throughout this work, we evaluate the released Ouro-1.4B and Ouro-2.6B checkpoints. We consider both fixed-depth inference, where every token executes a predetermined number of recurrent iterations, and adaptive inference using the pretrained halting gate.

\subsection{Retrofitted Recurrent Models}
\label{app:retrofit}

In addition to the native recurrent Ouro architecture, we evaluate the retrofitted recurrent language models of \citet{retrofittedrecurrence}, which convert pretrained decoder-only Transformers into looped models while largely preserving their pretrained parameters. The original Transformer is partitioned into three components: a non-recurrent prelude, a shared recurrent block, and a non-recurrent coda. The recurrent block consists of a contiguous subset of Transformer layers whose parameters are shared across recurrent iterations. During inference, the prelude is executed once, the recurrent block is repeatedly applied for a configurable number of iterations, and the coda is executed once before producing the next-token prediction. This retrofit increases inference-time computation through repeated latent refinement while maintaining the initialization and capabilities of the original pretrained model.

In contrast to Ouro, the retrofitted model does not learn an adaptive halting policy. Instead, the number of recurrent iterations $r$ is sampled independently for each training batch from a Poisson-lognormal depth distribution, \( \lambda \sim \mathrm{LogNormal}(\mu,\sigma^2) \text{, } r \sim \mathrm{Poisson}(\lambda), \) where $\mu$ and $\sigma$ control the expected recurrence depth. The model is then supervised only after the sampled number of recurrent iterations using the objective
\[ \mathcal{L}(\theta) = \mathbb{E}_{x} \mathbb{E}_{r\sim D_{\mathrm{PLN}}} \left[ \sum_{\ell=1}^{M-1} -\log p_{\theta}^{(r)} (x_{\ell+1}\mid x_{1:\ell}) + \beta D_{\mathrm{KL}} \!\left( p_{\theta_0}(\cdot\mid x_{1:\ell}) \;\|\; p_{\theta}^{(r)}(\cdot\mid x_{1:\ell}) \right) \right], \]
where $p_{\theta}^{(r)}$ denotes the next-token distribution after $r$ recurrent iterations and $p_{\theta_0}$ is the corresponding distribution of the frozen pre-adaptation model. The KL regularization preserves the behavior of the original pretrained model while adapting it to recurrent computation. Unlike Ouro, which jointly optimizes predictions across all recurrent iterations using a learned exit distribution, the retrofitted model computes the language-model loss only from the final recurrent prediction corresponding to the sampled recurrence depth.

Throughout this work, we evaluate released retrofitted recurrent checkpoints initialized from two pretrained model families: OLMo-2-0425-1B and Llama-3.2-1B. Since the model does not include a learned halting mechanism, we evaluate it only under fixed-depth inference by varying the number of recurrent iterations executed during generation.

\section{Hermes Function-Calling Dataset}
\label{app:hermes_dataset}

All supervised fine-tuning experiments described in this work use the Hermes Function-Calling V1 dataset introduced by Nous Research \citep{hermesfunctioncalling, teknium2024hermes3technicalreport}. The dataset is a synthetic instruction-following corpus designed for training language models to produce structured function calls and JSON outputs from natural-language requests. It combines single-function and multi-function tool-calling conversations together with structured extraction, JSON-mode, and agentic interaction examples, all formatted according to the Hermes Function-Calling standard. The training examples include user requests, tool definitions expressed through JSON schemas, assistant-generated tool calls, tool responses, and final assistant replies, allowing models to learn both API selection and argument generation in multi-turn settings. The released dataset also incorporates updated function calling data derived from the Glaive function-calling corpus and additional synthetic tool use examples created by Nous Research. Throughout this work, all controlled supervised fine-tuning experiments use the same Hermes Function-Calling V1 training split and ChatML formatting described in Section~\ref{sec:experimental_setup}.

\section{Additional Qualitative Examples}
\label{app:qualitative-examples}

This section provides further qualitative examples of how predictions change with
recurrent depth, complementing Figure~\ref{fig:qualitative_refinement}. All examples
are drawn from the NESTful subset described in Section~\ref{res:depth},
using fixed-depth inference. In each figure, the tool catalogue is the full candidate
set supplied to the model for that instance, and calls are shown in the compact form
\texttt{function(argument=value) $\rightarrow$ label}. The examples here illustrate the error types we observe at low recurrent depth: functions absent from
the supplied catalogue, argument names drawn from other tool specifications,
incorrect call ordering, spurious additional calls, and omitted dependent calls.

\begin{figure}[!t]
    \centering
    \includegraphics[width=0.6\textwidth]{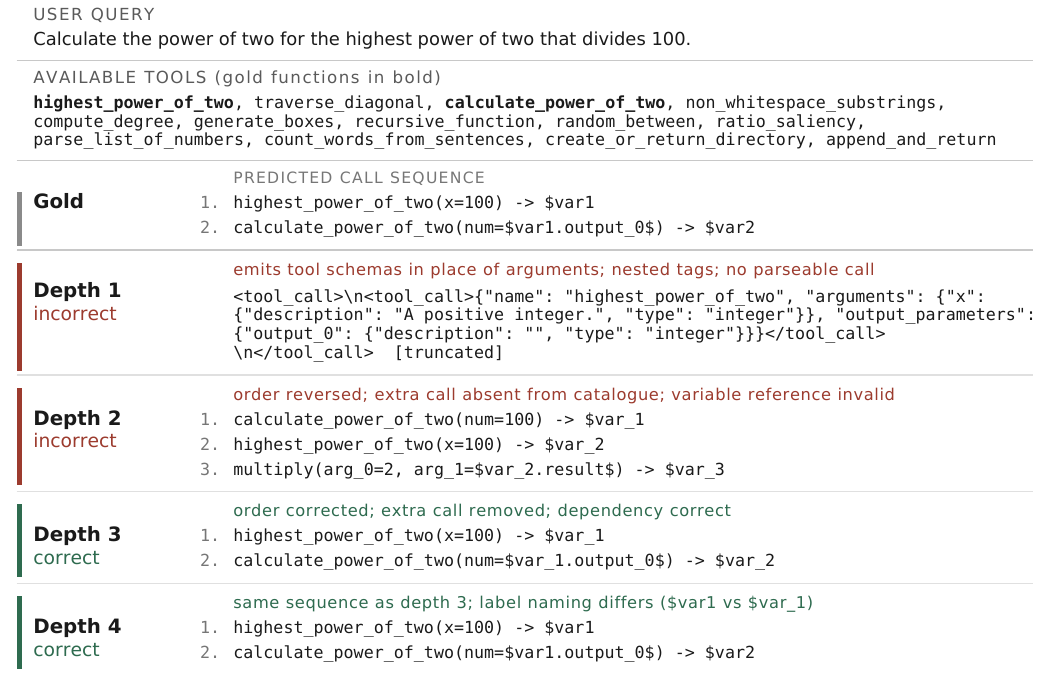}
    \caption{\textbf{Ordering and spurious-call correction.} At depth 1 the output is malformed, emitting tool specifications rather than call arguments (excerpt shown truncated). At depth 2 both gold functions appear but in reversed order, followed by a call to \texttt{multiply}, which is absent from the tool catalogue, and an invalid variable reference. Depths 3 and 4 match the gold sequence, differing from each other only in label naming.}
    \label{fig:qual_ordering}
\end{figure}

Figure~\ref{fig:qual_ordering} illustrates several failure modes being corrected over
recurrent refinement. The initial prediction is malformed, while the second iteration
produces a syntactically valid but semantically incorrect sequence that introduces an
unsupported function. Subsequent recurrent steps eliminate both the ordering error and
the hallucinated tool invocation, converging to the gold execution trace without
further modification.

\begin{figure}[!t]
    \centering
    \includegraphics[width=0.6\textwidth]{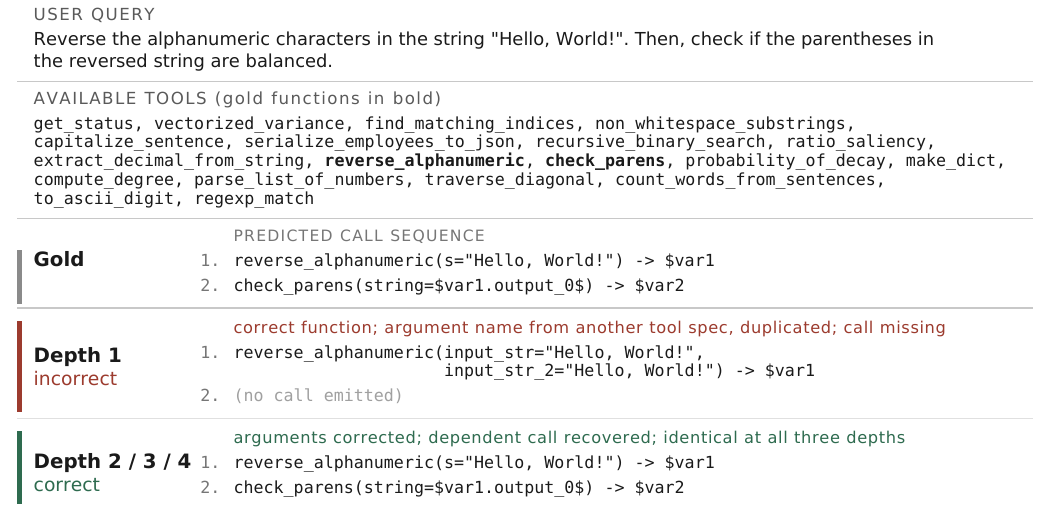}
    \caption{\textbf{Argument correction.} At depth 1 the model selects the correct first function but names its argument \texttt{input\_str}, a parameter of a different tool in the same catalogue, duplicates it as \texttt{input\_str\_2}, and omits the dependent call. Depths 2 through 4 match the gold sequence and are identical to one another.}
    \label{fig:qual_arguments}
\end{figure}

The remaining examples isolate narrower changes. Figure~\ref{fig:qual_arguments}
shows a case where function selection is already correct at depth 1 and only the
argument names differ from the specification. The model supplies \texttt{input\_str},
which is a parameter of a different tool in the same catalogue, and repeats it as a
second argument. The correction at depth 2 therefore affects argument grounding and
the presence of the dependent call, rather than the choice of function.

\begin{figure}[!t]
    \centering
    \includegraphics[width=0.6\textwidth]{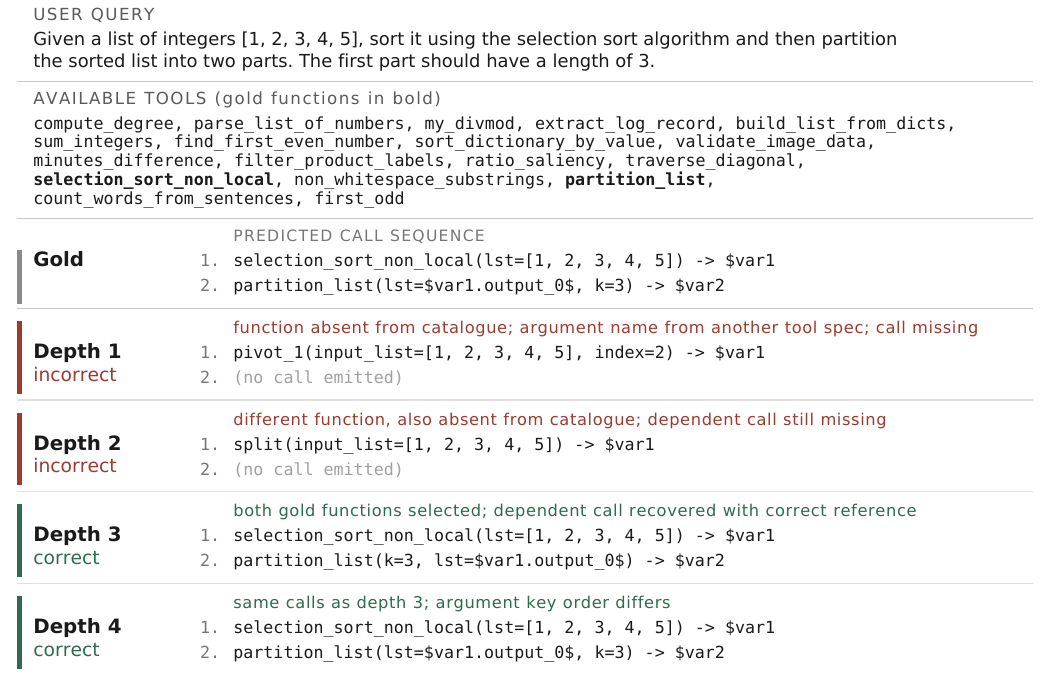}
    \caption{\textbf{Two distinct incorrect attempts before recovery.} Depths 1 and 2 each emit a single call to a function absent from the tool catalogue and omit the required dependent call. At depth 1, the model also uses an argument name belonging to a different tool in the catalogue. Depths 3 and 4 select both gold functions and correctly pass \texttt{\$var1.output\_0\$}, differing only in argument-key order.}
    \label{fig:qual_two_attempts}
\end{figure}

Figure~\ref{fig:qual_two_attempts} shows a case in which the prediction changes
between two consecutive incorrect depths. Depths 1 and 2 name different functions,
neither of which appears in the supplied catalogue, and both omit the second call.
The two correct depths differ only in the order of the keys within the second call's
argument dictionary, which the evaluator treats as equivalent.

\begin{figure}[!t]
    \centering
    \includegraphics[width=0.6\textwidth]{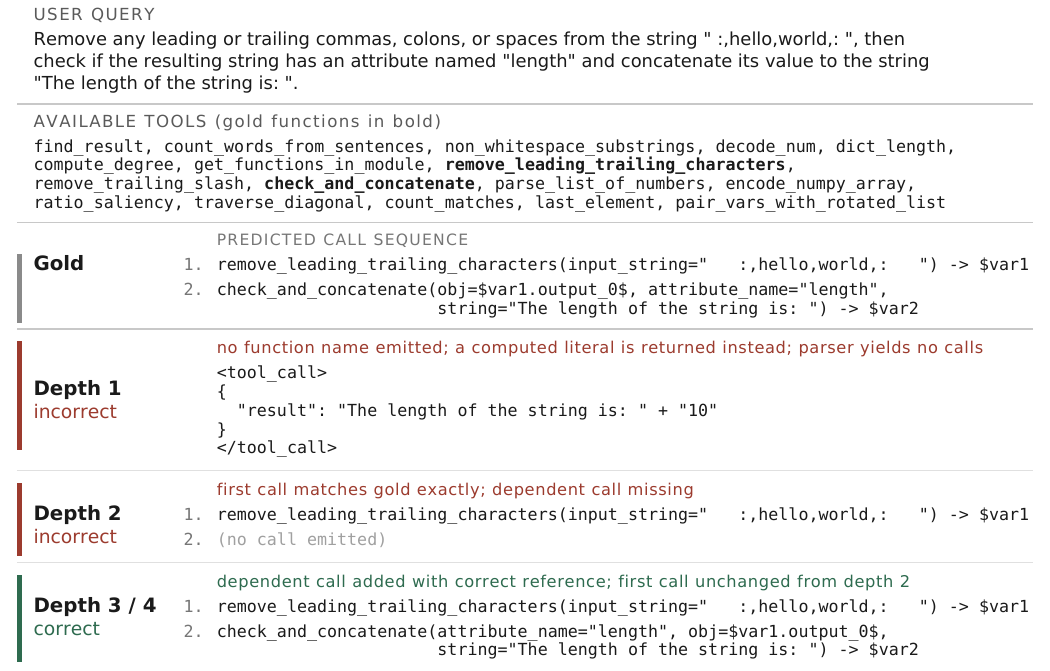}
    \caption{\textbf{Isolated recovery of a missing dependent call.} Depth 1 emits an object with no function name, returning a computed literal, so the parser yields no calls. Depth 2 emits a first call that matches the gold sequence exactly but stops there. Depths 3 and 4 leave that first call unchanged and add the dependent call with the reference \texttt{\$var1.output\_0\$}, isolating the recovery of the dependency from any change in function selection.}
    \label{fig:qual_dependency}
\end{figure}

Finally, Figure~\ref{fig:qual_dependency} separates the recovery of a dependency from any change in function selection. The first call is already identical to the gold call at depth 2, and the only difference between depths 2 and 3 is the addition of the second call together with the reference \texttt{\$var1.output\_0\$}. Depth 1 in this example produces an object with no function name at all, which the parser does not resolve to a call.

Across these examples, the changes observed with additional recurrent depth involve which functions are named, which arguments they receive, how many calls are emitted, and how intermediate outputs are referenced. We describe only observable differences between predictions and the gold sequence, and we do not claim that these figures are representative of the error distribution over the full evaluation set.

\section{Tool-Calling Benchmark Examples}
\label{app:benchmark_examples}

This section provides representative examples from the tool-calling
benchmarks evaluated in this work and relates their task structures to the
formalism introduced in Section~\ref{sec:background_tool_calling}. We represent a tool use solution as $G_x=(C_x,E_x)$, where $C_x$ is the set of required function calls and $E_x$ contains output-to-input dependencies between calls. The benchmarks emphasize complementary aspects of this structure: API-Bank focuses on evaluating the grounding of individual calls, BFCL evaluates function selection and the construction of independent multi-call sets, and NESTful evaluates dependent call graphs in which intermediate outputs are consumed by subsequent calls. Table~\ref{tab:benchmark_comparison} summarizes these differences before we present representative examples from each benchmark.

\begin{table}[!htpb]
\centering
\caption{Structural comparison of the evaluated tool-calling benchmarks under the solution representation $G_x=(C_x,E_x)$. API-Bank emphasizes individual-call grounding, BFCL evaluates increasingly complex independent call sets, and NESTful introduces explicit output-to-input dependencies.}
\label{tab:benchmark_comparison}

\footnotesize
\setlength{\tabcolsep}{5pt}
\renewcommand{\arraystretch}{1.12}
\begin{tabular}{lcccc}
\toprule
Benchmark
& Candidate tools
& Call structure
& Dependencies
& Primary challenge \\
\midrule
API-Bank
& one or more
& typically $|C_x|=1$
& $E_x=\varnothing$
& Tool and argument grounding \\

BFCL Simple
& $|\mathcal{T}_x|=1$
& $|C_x|=1$
& $E_x=\varnothing$
& Argument grounding \\

BFCL Multiple
& $|\mathcal{T}_x|>1$
& $|C_x|=1$
& $E_x=\varnothing$
& Function selection \\

BFCL Parallel
& one or more
& $|C_x|>1$
& $E_x=\varnothing$
& Independent call generation \\

BFCL Parallel-Multiple
& $|\mathcal{T}_x|>1$
& $|C_x|>1$
& $E_x=\varnothing$
& Selection and call composition \\

NESTful
& multiple
& $|C_x|>1$
& typically $|E_x|>0$
& Dependency-aware execution \\
\bottomrule
\end{tabular}
\end{table}

\subsection{Berkeley Function Calling Leaderboard}
\label{app:bfcl_examples}

The Berkeley Function Calling Leaderboard (BFCL) evaluates whether a language model can select and instantiate functions from natural-language requests \citep{bfcl}. We evaluate the non-live single-turn categories: Simple, Multiple, Parallel, and Parallel-Multiple. These categories differ in the number of candidate tools supplied to the model and in the number of calls that must be generated. They do not require the output of one call to be consumed by another. Such output-to-input dependencies are instead studied by the NESTful dataset.

Using the notation introduced in Section~\ref{sec:background_tool_calling}, let $\mathcal{T}_x \subseteq \mathcal{T}$ denote the candidate tools supplied with request $x$, and let \( C_x = \{c_1,\ldots,c_K\} \text{, } c_k=(f_k,a_k), \) denote the reference call set. For the BFCL single-turn categories considered here, the calls within an example are independent, and hence the dependency graph satisfies $E_x=\varnothing$. The categories differ mainly in $|\mathcal{T}_x|$ and $|C_x|$. The following examples are reproduced in shortened form from the executable counterparts of the corresponding BFCL categories. Tool descriptions are shortened for readability.

\paragraph{Simple.}
Simple examples supply a single tool and require one call: \( |\mathcal{T}_x|=1, \text{ } |C_x|=1, \text{ } E_x=\varnothing. \) For example, the request

\begin{quote}
\small
``A biased die produces a six with probability $0.6$. If it is rolled
$20$ times, what is the probability of obtaining exactly five sixes?''
\end{quote}

is accompanied by the tool \( \texttt{calc\_binomial\_probability(n, k, p)}, \) and has the reference call \( C_x = \left\{ \texttt{calc\_binomial\_probability(n=20, k=5, p=0.6)} \right\}. \) This category tests argument extraction and schema-conformant call
generation.

\paragraph{Multiple.}
Multiple examples supply between two and four candidate tools but require only
one call: \( |\mathcal{T}_x|>1, \text{ } |C_x|=1, \text{ } E_x=\varnothing. \)
For example, a request asks for the probability of obtaining exactly 5
sixes in $20$ fair-die rolls. The candidate set contains both
\( \texttt{get\_weather\_data(coordinates)} \text{ and } \texttt{calc\_binomial\_probability(n, k, p)}. \)
The correct call is \( C_x = \left\{ \texttt{calc\_binomial\_probability(n=20, k=5, p=1/6)} \right\}. \)
Thus, ``Multiple'' refers to the presence of multiple candidate function
definitions, rather than to the generation of multiple calls. The category
tests function selection in addition to argument grounding.

\paragraph{Parallel.}
Parallel examples require multiple independent invocations of a supplied
function: \( |C_x|>1, \text{ } E_x=\varnothing. \) For example, BFCL includes the request

\begin{quote}
\small
``Play songs from Taylor Swift and Maroon 5 for 20 minutes and
15 minutes, respectively, on Spotify.''
\end{quote}

with the tool \( \texttt{spotify.play(artist, duration)}. \) The reference call set is
\[
C_x =
\left\{
\begin{aligned}
&\texttt{spotify.play(artist=\{Taylor Swift\}, duration=20)},\\
&\texttt{spotify.play(artist=\{Maroon 5\}, duration=15)}
\end{aligned}
\right\}.
\]
The calls do not consume one another's outputs and may therefore be executed
concurrently. The model must identify the correct number of calls and align
each entity with its corresponding arguments.

\paragraph{Parallel-Multiple.}
Parallel-Multiple combines multiple candidate functions with the generation
of multiple independent calls: \( |\mathcal{T}_x|>1, \text{ } |C_x|>1, \text{ } E_x=\varnothing. \) For example, a request asks both for the current weather in Ottawa and for the probability of obtaining 5 wins in ten independent attempts with success probability $0.5$. The candidate tools are \( \texttt{get\_weather\_data(coordinates)}
\text{ and }
\texttt{calc\_binomial\_probability(n, k, p)}, \)
and the reference call set is
\[
C_x =
\left\{
\begin{aligned}
&\texttt{get\_weather\_data(
    coordinates=[45.4215,-75.6972])},\\
&\texttt{calc\_binomial\_probability(
    n=10, k=5, p=0.5)}
\end{aligned}
\right\}.
\]
This category jointly tests function selection, call-set construction, and
argument assignment across different tools.

\subsection{NESTful}
\label{app:nestful_examples}

NESTful evaluates nested sequences of executable API calls in which the output of one function is passed as an argument to a subsequent function \citep{nestful}. Each instance contains a user request, a catalog of available tools, a gold sequence of calls with arguments, and the final answer obtained by executing that sequence. The released evaluation set contains 1861 instances drawn from mathematical reasoning and coding domains, together with executable implementations of the corresponding functions.

Using the notation of Section~\ref{sec:background_tool_calling}, a NESTful solution is represented as \( G_x=(C_x,E_x), \) where \( C_x=\{c_1,\ldots,c_K\} \) contains the required function calls and \( (c_i,c_j)\in E_x \) whenever the output of $c_i$ is used to instantiate an argument of $c_j$. Unlike the BFCL single-turn categories considered in this work, NESTful therefore evaluates both call-set construction and explicit output-to-input dependencies. Its call structure is generally a directed acyclic graph rather than merely a collection of independent calls. NESTful assigns a unique label to the output of every call. If a call $c_i$ is assigned the label \texttt{\$var\_i}, a later argument can refer to one of its output fields using notation such as \( \texttt{\$var\_i.result\$}. \) This makes the dependency relation explicit in the serialized reference sequence. For example, consider the request
\begin{quote}
\small
``Find the average of all the numbers between 6 and 34 that are divisible by 5.''
\end{quote}

The corresponding solution consists of the call set \( C_x = \{c_1,c_2,c_3,c_4\}, \) where
\[
\begin{aligned}
c_1 &= \texttt{add}(6,4),\\
c_2 &= \texttt{subtract}(34,4),\\
c_3 &= \texttt{add}\!\left(\operatorname{out}(c_1),
                           \operatorname{out}(c_2)\right),\\
c_4 &= \texttt{divide}\!\left(\operatorname{out}(c_3),2\right).
\end{aligned}
\]

The corresponding dependency graph has \( C_x=\{c_1,c_2,c_3,c_4\} \) and \( E_x=\{(c_1,c_3),(c_2,c_3),(c_3,c_4)\}. \) Calls $c_1$ and $c_2$ are independent and may be evaluated in parallel. Call $c_3$ depends on both of their outputs, while $c_4$ depends on the output of $c_3$. The longest dependency path therefore has two edges:
\[
c_1 \rightarrow c_3 \rightarrow c_4
\qquad\text{or}\qquad
c_2 \rightarrow c_3 \rightarrow c_4.
\]
Executing the complete sequence produces the gold answer $20$.

This example illustrates the distinction between node composition and
dependency composition. The model must first select the correct calls
and instantiate their constant arguments, as in ordinary multi-call
generation. It must additionally assign outputs to variables, select the
correct output fields, and bind those variables to the arguments of subsequent
calls. Errors can therefore arise from selecting an incorrect function,
constructing an incorrect argument, omitting a call, producing the wrong
ordering, or referencing the wrong intermediate variable. NESTful reports complementary metrics at different levels of the predicted solution. Function and parameter scores measure the correctness of individual calls, partial sequence matching measures how much of the gold call sequence is recovered, and full sequence matching requires the complete sequence of functions and arguments to match. The benchmark additionally executes the predicted calls and reports a win rate based on whether the resulting answer is correct. These metrics distinguish locally plausible calls from complete, dependency-preserving solutions.

\subsection{API-Bank}
\label{app:apibank_examples}

API-Bank evaluates tool-augmented language models through conversational API usage \citep{apibank}. Each instance consists of a user request, a collection of available APIs, an expected API invocation, the API response, and the final assistant response. Compared to BFCL and NESTful, API-Bank evaluates the grounding of individual tool calls rather than the construction of multi-call workflows. An API-Bank example is represented as \( G_x=(C_x,E_x), \) where \( |C_x|=1,\text{ } E_x=\varnothing. \) The task is therefore to select the correct function and instantiate its arguments from the user request before integrating the returned observation into the final response.

For example, consider the dialogue ``Can you calculate $(5+6)\times 3$ for me?'' The available API is \( \texttt{Calculator(formula)}, \) where \texttt{formula} is a string containing an arithmetic expression. The corresponding reference call is \( C_x = \left\{ \texttt{Calculator}\!\left( \texttt{formula}=\texttt{'(5+6)*3'} \right) \right\}. \) Since the instance requires a single API invocation, its dependency relation is \( E_x=\varnothing. \)

The example tests whether the model can identify the required API, translate the natural-language request into the expected argument representation, and produce a correctly formatted invocation. More generally, API-Bank evaluates API selection and argument generation within conversational contexts, which may contain information accumulated across multiple dialogue turns. Unlike BFCL, which explicitly evaluates the construction of independent multi-call sets, and NESTful, which evaluates output-to-input dependencies between calls, the API-Bank setting considered in our evaluation emphasizes the grounding and invocation of individual APIs.

\end{document}